\documentclass[letterpaper, 10 pt, conference]{ieeeconf}
\IEEEoverridecommandlockouts
\usepackage{cite}

\usepackage{amsmath,amssymb,amsfonts}
\usepackage{algorithmic}
\usepackage{graphicx}
\usepackage{textcomp}

\usepackage{xcolor}
\def\BibTeX{{\rm B\kern-.05em{\sc i\kern-.025em b}\kern-.08em
    T\kern-.1667em\lower.7ex\hbox{E}\kern-.125emX}}
\begin{document}

\title{\bf
Online Estimation of Table-Top Grown Strawberry Mass in Field Conditions with Occlusions
}

\author{Jinshan Zhen†$^1{^{,2}}$, Yuanyue Ge†$^1$, Tiaoxiao Zhu†$^1{^{,3}}$, Hui Zhao$^2$ and Ya Xiong$^1$* 
\thanks{This work was supported by the Haidian District Bureau of Agriculture and Rural Affairs, the Innovation Ability Project of Beijing Academy of Agricultural and Forestry Sciences (BAAFS) (KJCX20240321), the Outstanding Youth Foundation of BAAFS (YKPY2025007), the Postdoctoral Research Fund of BAAFS, the BAAFS Talent Recruitment Program, and the NSFC Excellent Young Scientists Fund (overseas). (*\textit{Corresponding Author: Ya Xiong, \tt\small yaxiong@nercita.org.cn})}
\thanks{†Jinshan Zhen, Yuanyue Ge and Tiaoxiao Zhu contributed equally to this work and are co-first authors.}
\thanks{$^1$The Intelligent Equipment Research Center, Beijing Academy of Agriculture and Forestry Sciences, Beijing 100097, China.} %
\thanks{$^2$The College of Electrical Engineering and Automation, Tianjin University of Technology, Tianjin, 300382, China.}%
\thanks{$^3$The College of Mechatronic Engineering and Automation, Shanghai  University, Shanghai, 200444, China.}%
}

\maketitle

\begin{abstract}
Accurate mass estimation of table-top grown strawberries under field conditions remains challenging due to frequent occlusions and pose variations. 
This study proposes a vision-based pipeline integrating RGB-D sensing and deep learning to enable non-destructive, real-time and online mass estimation. The method employed YOLOv8-Seg for instance segmentation, Cycle-consistent generative adversarial network (CycleGAN) for occluded region completion, and tilt-angle correction to refine frontal projection area calculations. A polynomial regression model then mapped the geometric features to mass.
Experiments demonstrated mean mass estimation errors of 8.11\% for not-occluded strawberries and 10.47\% for occluded cases. CycleGAN outperformed large mask inpainting (LaMa) model in occlusion recovery, achieving superior pixel area ratios (PAR) (mean: 0.978 vs. 1.112) and higher intersection over union (IoU) scores (92.3\% vs. 47.7\% in the [0.9–1] range). This approach addresses critical limitations of traditional methods, offering a robust solution for automated harvesting and yield monitoring with complex occlusion patterns.
\end{abstract}



\section{INTRODUCTION}

Fruit mass estimation is essential for optimizing harvest timing, improving agricultural efficiency, and advancing smart, precision agriculture \cite{b1}. Additionally, fruit size and mass serve as key criteria for grading, aligning with the increasingly diverse demands of the commercial market \cite{b2}. However, traditional electronic scale-based methods can damage the delicate skins of fruits, leading to economic losses, while manual grading requires substantial human labor. 
Mass estimation for strawberries poses even greater challenges, especially in strawberry-harvesting robots, where the mass estimation must be conducted during the picking process, thus imposing higher demands on real-time performance and non-destructive operation.
In field conditions, strawberries are often occluded by stems, leaves, or other fruits, complicating fruit modeling and causing significant deviations between the estimated and actual mass values \cite{xiong2020autonomous}. 

Currently, the primary non-destructive measurement methods for fruit mass include near-infrared spectroscopy \cite{b011}, laser scanning \cite{b012}, and computer vision image processing \cite{b013}. In general, computer vision methods offer significant advantages in terms of efficiency, multi-dimensional information acquisition, automation, and robotization, making them more widely applicable and convenient.

In terms of shape recovery for fruits under occluded conditions, circular, elliptical, and other regular-shaped fruits are typically easier to reconstruct in terms of shape under occluded conditions. 
For example, in \cite{b13}, an apple size estimation method combining an RGB-D camera with the geometric shape of the apple was introduced. This model employed binary masks and shape fitting algorithms to address occlusion issues. Alternatively, \cite{b16} utilized Retanet to detect melons in images, applying the Chan-Vese active contour algorithm and PCA ellipse fitting method to extract geometric features from the detected cantaloupe images. A regression model was then developed based on these geometric features, linking the elliptical characteristics (such as the major and minor axes) with melon mass to achieve accurate mass estimation. 

For irregular-shaped fruits, such as strawberries, mass estimation is more challenging using traditional geometry-based shape recovery methods. Existing methods have primarily focused on the mass estimation of not-occluded strawberries. Tafuro et al. introduced a mass estimation method that integrated RGB-D data, leveraging principal component analysis (PCA) for dimensionality reduction and a Random Forest model for mass prediction \cite{tafuro2022strawberry}. Similarly, Huang et al. developed a strawberry weight estimation method using plane-constrained binary division point cloud completion \cite{huang2024strawberry}. This approach required minimal data and achieves up to a 20.95\% improvement in prediction accuracy compared to state-of-the-art methods. Additionally, Basak et al. explored a strawberry weight estimation technique based on machine learning, employing linear regression (LR) and support vector regression (SVR) models to predict weight while analyzing the comparative advantages of both approaches \cite{basak2022non}. In contrast to the above methods that only handle non-occluded strawberries, the method proposed in \cite{b016} addresses calyx occlusion using right kite and simple kite geometry models, but it is limited in recovering or reconstructing other parts of the strawberry. Also, Ge et al. proposed a symmetry-plane-based method to complete the strawberry shape in 3D, but it struggled to handle deformed point clouds from RGB-D cameras \cite{Ge_symmetry}.

\begin{figure*}[t!]
    \centering
    \includegraphics[width=18cm]{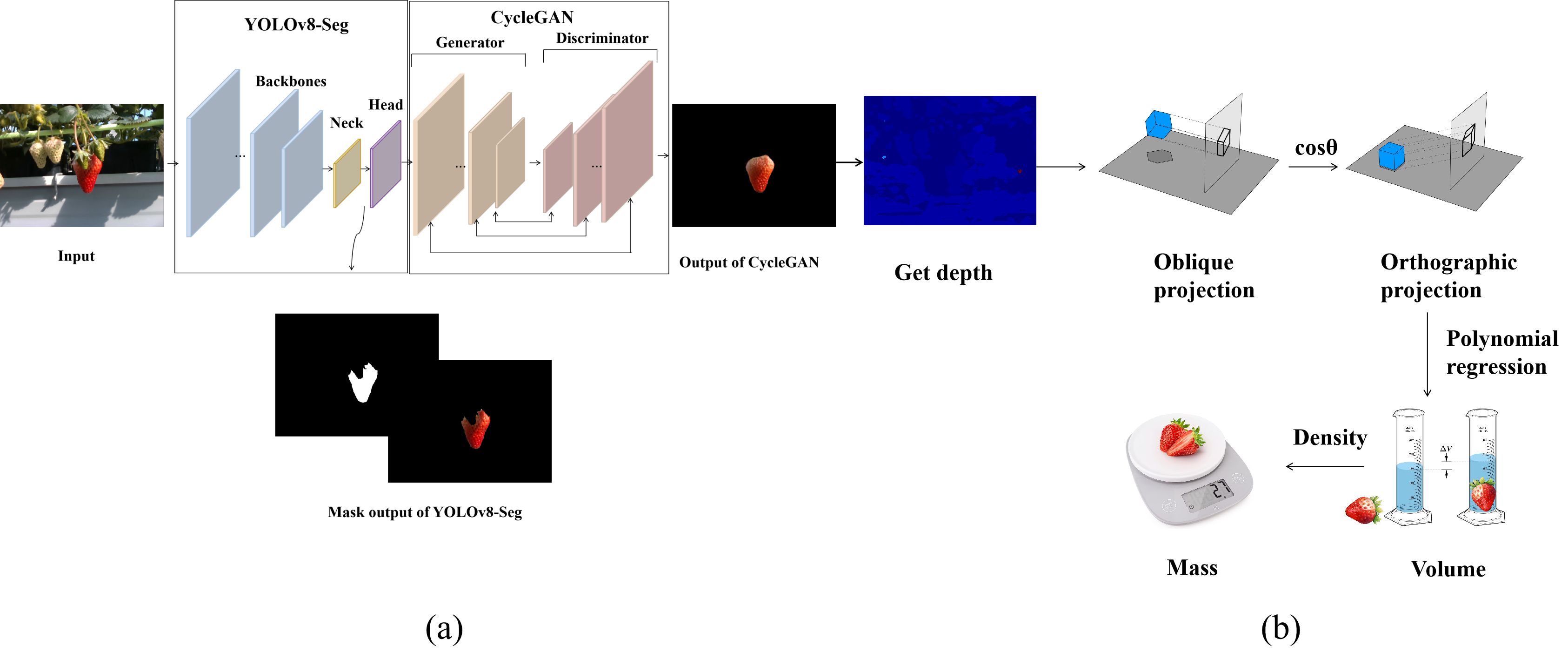}
    \caption{\textbf{Pipeline of mass estimation for occluded strawberries:} 
(a) The RGB image is first processed by an instance segmentation network to isolate the strawberry region, followed by occlusion restoration using CycleGAN; 
(b) The restored strawberry image is projected into a 3D space using a depth map and camera geometry for volume estimation, and the mass is subsequently predicted based on a volume-to-mass regression model derived from empirical strawberry density measurements.}
    \label{fig:flow chart}
\end{figure*}


Generative Adversarial Networks (GANs) are deep learning-based models consisting of a Generator and a Discriminator, which are optimized through adversarial training to generate more realistic samples. GANs have been widely applied in image generation, style transformation, data augmentation, and image completion, achieving significant advancements in these areas. Although several GAN-based image completion methods have been proposed \cite{b22, b23, b24}, they often suffer from training instability and are primarily designed for local inpainting, which limits their ability to reconstruct large missing regions.

ShapeMorph\cite{10944053} realizes high-precision and diverse 3D completion with excellent detail fidelity through irregular coding and chunked diffusion. However, it has two major drawbacks: slow inference speed and the inability to semantically control the generated results. The diffusion model\cite{chu2023diffcompletediffusionbasedgenerative3d} performs well in improving the accuracy and generalization ability of 3D shape complementation, but still has two major drawbacks: first, the inference process is multi-step iterative and has high computational overhead, which makes it difficult to meet real-time demands; second, the control ability is limited, and there is a lack of explicit semantic regulation of the generated shapes, which makes it difficult to realize controllable generation and interactive editing.

CycleGAN \cite{b8x} offers notable advantages due to its unsupervised learning capability, cross-domain image translation, and cycle consistency. Leveraging these strengths, we applied CycleGAN to recover occluded regions of strawberries, enabling mass estimation as if the fruit were fully visible. To further enhance accuracy, we integrated the strawberry mass estimation pipeline with CycleGAN-based occlusion recovery, thereby improving the robustness and adaptability of the method in complex field environments.

Our contributions are as follows: 
\begin{itemize}

\item[$\bullet$] We proposed an image restoration method to complete the occluded parts of strawberries, providing an accurate solution for strawberry mass estimation in complex field environments. 

\item[$\bullet$]  An online visual strawberry mass estimation pipeline was introduced, enabling direct mass determination using an RGB-D camera that accounted for pose variations. 


\end{itemize}




\section{METHODS}

Fig.~\ref{fig:flow chart} shows the entire pipeline of the proposed strawberry mass estimation method. First, YOLOv8-Seg was used for instance segmentation to achieve accurate separation of strawberry pixels from the background. For the occlusion case, CycleGAN adaptively reconstructed the occluded region. Subsequently, the RGB image and depth information were combined, and the projection angle was corrected to accurately obtain the cross-sectional area of the strawberry. Finally, volume and mass were calculated by regression analysis to achieve high-precision mass prediction.

\subsection{Strawberry segmentation}

Although various instance segmentation models, such as Mask R-CNN \cite{b017} and U-Net \cite{b018}, have been developed, they fail to meet the real-time processing demands of our application. To address this limitation, we adopted YOLOv8-Seg, which offers superior real-time performance, as our instance segmentation solution. YOLOv8-Seg leverages the enhanced CSPDarknet53 \cite{b019} as its backbone network and incorporates a Dynamic Mask Head to achieve precise, pixel-level segmentation. By building upon the strengths of its predecessors, YOLOv8-Seg further enhanced instance segmentation by directly extracting both the Mask and the Region of Interest (RoI). This advancement significantly improved segmentation accuracy and efficiency while ensuring robust recognition performance, even in the presence of overlapping or occluded objects.

In the data acquisition phase, we used an Intel RealSense D435 RGB-D camera to capture 1200 high-resolution strawberry images from the strawberry tunnels. The dataset was divided into training, testing, and validation sets in a 6:3:1 ratio, encompassing a variety of lighting conditions and varying degrees of occlusion. This approach enhanced the model’s generalization ability and ensured data diversity and complexity. Using the Labelme annotation platform, we meticulously annotated the images with a polygonal segmentation tool to precisely outline the strawberry instances. The samples were then classified into two categories: occluded and isolated. This categorization facilitated the subsequent restoration process, where the shape of the strawberry could be better understood and reconstructed.

\subsection{Shape completion for occluded strawberries}

CycleGAN is an unsupervised deep learning model for image-to-image translation, capable of transforming images between different domains without requiring paired training data. In the task of strawberry occlusion recovery, we leveraged this capability by dividing the image into two domains: one representing the occluded image and the other representing the non-occluded image. These two domains were derived from the output of the YOLOv8 instance segmentation model.

\begin{figure}[htp]
    \centering
    \includegraphics[width=9cm]{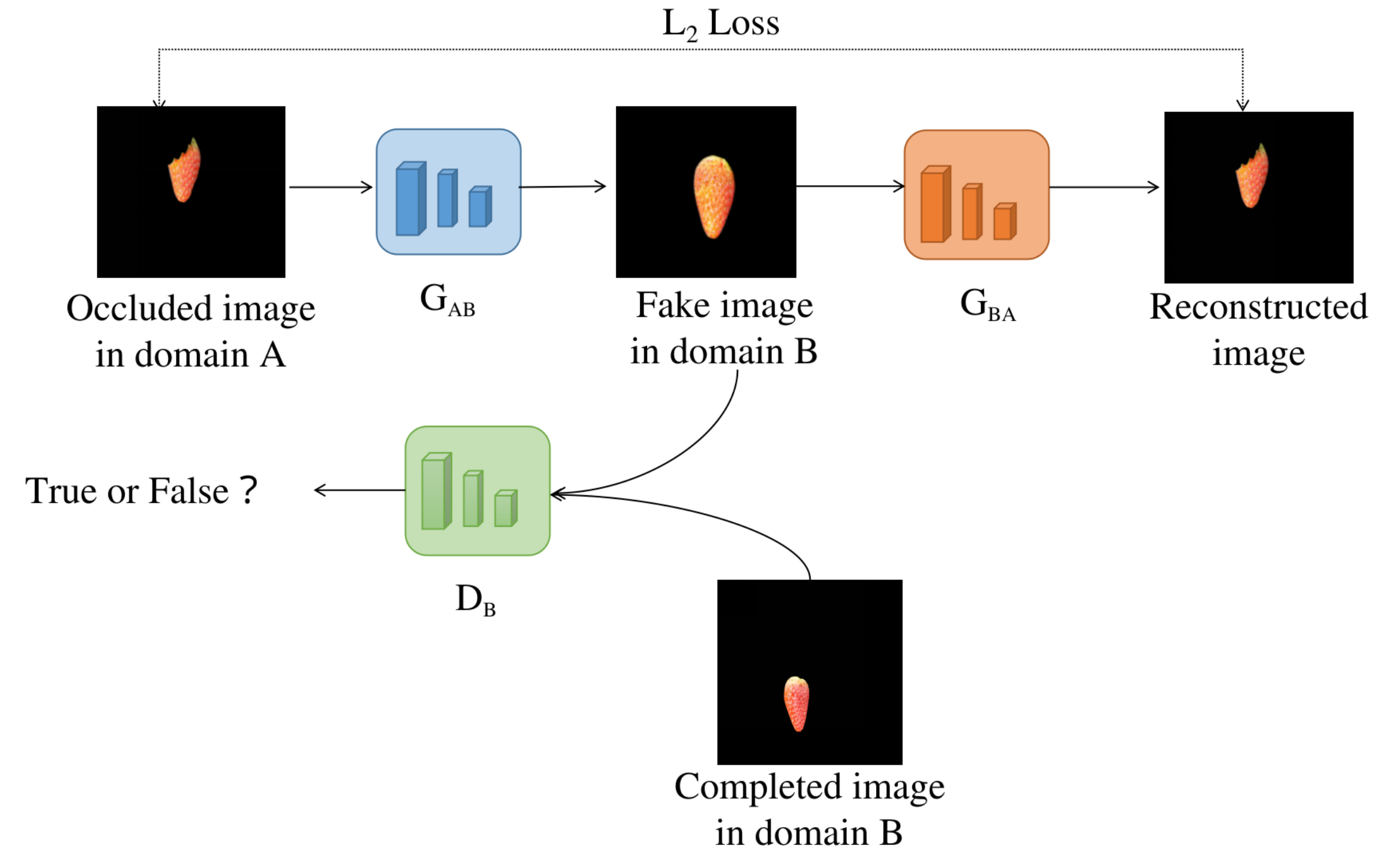}
    \caption{\textbf{CycleGAN-based shape completion framework:} The generator $G_{AB}$ transforms an occluded strawberry image from domain A into a completed (fake) image in domain B, which is then passed to the discriminator $D_B$ for adversarial training against real images in domain B. Meanwhile, the reverse generator $G_{BA}$ reconstructs the original occluded image, and an $L_2$ loss is applied between the input and reconstructed image to enforce cycle consistency.}
    \label{fig:gan}
\end{figure}

The working schematic of CycleGAN is shown in Fig.\ref{fig:gan}. Specifically, the strawberry images were divided into occlusion domain A and completeness domain B. CycleGAN generated plausible content for the occluded regions through bi-directional mapping by converting the occluded image (domain A) into a completed image (domain B) and mapping it back from domain B to domain A, with a cycle consistency loss to ensure alignment with the original image in terms of visual features and semantic structure. In the network structure, the generator G\textsubscript{AB} was responsible for the conversion from occluded image to completed image, and the generator G\textsubscript{BA} reconstructed the occluded version from the completed image. Both domain A (occluded images) and domain B (complete images) consist of real data naturally collected from field environments. The conversion from domain A to domain B is accomplished through adversarial training between the generator and discriminator, as defined in the CycleGAN framework.

The final optimization objective of CycleGAN was obtained by the weighted sum of the cycle consistency loss and two adversarial losses. Among them, the cycle consistency loss ensured that the image could be recovered to its original form after bi-directional conversion, thus maintaining structural and content consistency. The adversarial loss matched the distribution of the generated image to that of the real image in the target domain, enhancing the realism of the conversion. The specific loss function formulas are shown below:


1) In the CycleGAN structure, the cycle consistency loss is defined as follows:
\begin{multline}
\mathcal{L}_{\mathrm{cyc}}(G, F)=\mathbb{E}_{x \sim p_{\text {data }}(x)}\left[\|F(G(x))-x\|_{1}\right] \\
+\mathbb{E}_{y \sim p_{\text {data }}(y)}\left[\|G(F(y))-y\|_{1}\right]
\end{multline}

where $G$ and $F$ represent the two generators. The sample $x$ is first mapped to domain $B$ by $G$, then mapped back to domain $A$ by $F$, ensuring it remains as close as possible to the original $x$. $\| \cdot \|_1$ represents the L1 loss, which is used to measure the pixel-wise difference before and after transformation, helping to preserve the original information in the generated image.

2) Two adversarial losses were employed for the unsupervised mappings $A \rightarrow B$ (via $G$) and $B \rightarrow A$ (via $F$). Together with the cycle consistency loss, they help ensure the realism, stability, and reversibility of domain translation. The two adversarial losses are defined as follows:

\begin{multline}
\mathcal{L}_{\text{GAN}}(G, D_Y, X, Y) = \mathbb{E}_{y \sim p_{\text{data}}(y)} \left[ \log D_Y(y) \right] \\
+ \mathbb{E}_{x \sim p_{\text{data}}(x)} \left[ \log (1 - D_Y(G(x))) \right]
\end{multline}

\begin{multline}
\mathcal{L}_{\text{GAN}}(F, D_X, Y, X) = \mathbb{E}_{x \sim p_{\text{data}}(x)} \left[ \log D_X(x) \right] \\
+ \mathbb{E}_{y \sim p_{\text{data}}(y)} \left[ \log (1 - D_X(F(y))) \right]
\end{multline}

where $x$ and $y$ denote the occluded and non-occluded strawberry images, respectively, and \( p_{\text{data}}(x) \) and \( p_{\text{data}}(y) \) represent their true distributions.

3) The final optimization objective of CycleGAN is defined as the weighted sum of the adversarial losses and the cycle consistency loss:
\begin{multline}
\mathcal{L}(G, F, D_X, D_Y) = \mathcal{L}_{\mathrm{GAN}}(G, D_Y, X, Y) \\
+ \mathcal{L}_{\mathrm{GAN}}(F, D_X, Y, X) + \lambda \mathcal{L}_{\mathrm{cyc}}(G, F)
\end{multline}

where the hyperparameter $\lambda$ controls the weight of the cycle consistency loss in the total loss, balancing adversarial learning and structural fidelity.

To enhance the diversity and robustness of the data, we applied data augmentation techniques, such as flipping, scaling, and panning, to the images. During the training process, we employed a staged strategy, dividing the generator's training into multiple phases, each focused on developing different capabilities of the generator and gradually guiding it to generate more complex images. Simultaneously, we increased the complexity of the discriminator in parallel with the generator's progress, ensuring that the discriminator could effectively evaluate the generator's output at each stage.

Additionally, we carefully tuned the training hyperparameters: the batch size was set to 2, and the number of convolutional kernels for both the generator and the discriminator was set to 96. The learning rate was set to 0.0002 for the first 100 epochs, then linearly decayed to 0.0001 for the subsequent 100 epochs.

Through these training strategies, CycleGAN successfully learned the mapping relationship between the two domains, enabling an efficient transformation from occlusion-broken images to intact images.

\subsection{Volume regression considering strawberry poses and mass estimation}

\begin{figure}[htp]
    \centering
    \includegraphics[width=7cm]{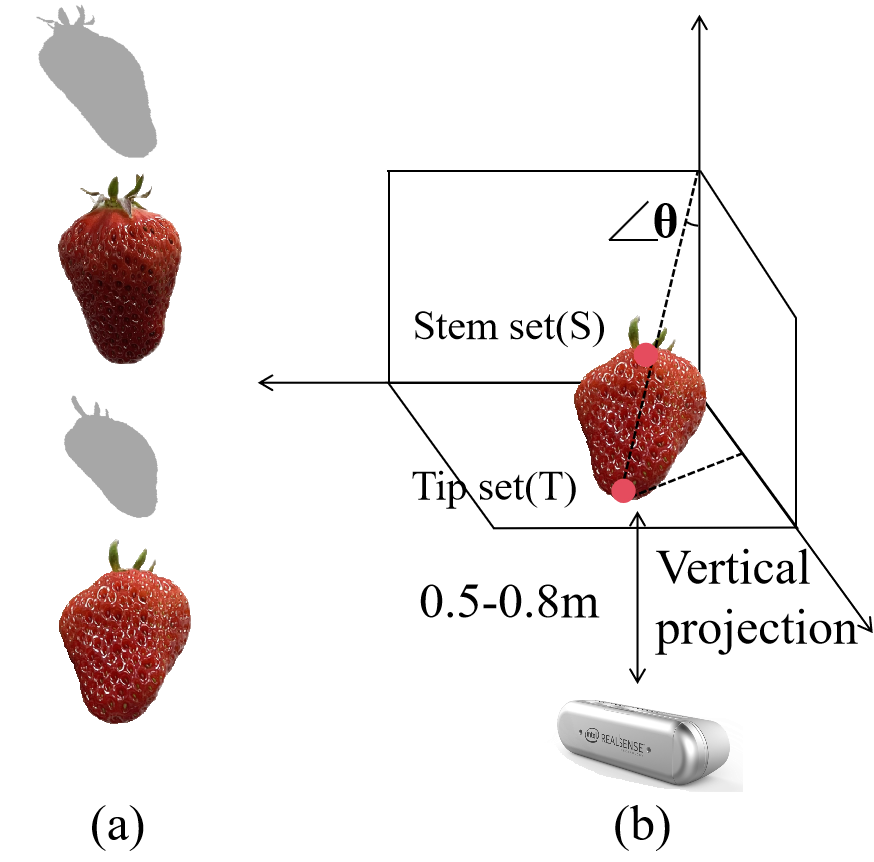}
    \caption{\textbf{Pose and projection of strawberries:} (a) Comparison of the projected visible areas between a vertically aligned strawberry and a tilted strawberry, highlighting how pose affects the observed contour. (b) Illustration of the tilt angle $\theta$ formed between the strawberry’s central axis and the vertical viewing direction, which is used for geometric correction in volume estimation.}
    \label{fig:pose}
\end{figure}

The visible area of the strawberry from the camera's perspective is usually smaller than the true frontal projected area because strawberries do not grow perfectly vertically. As shown in (a) of Fig.~\ref{fig:pose}, the visible area of a naturally grown tilted strawberry is smaller than that of a vertically oriented strawberry. This highlights the necessity of estimating the strawberry's pose. By leveraging the symmetry of the strawberry’s shape, we can obtain the tilt angle and correct the projection error using $\cos\theta$. Therefore, it is necessary to map the true frontal projection from the oblique angle, which requires estimating the pose of the strawberry. 

To achieve this, we proposed a method for estimating the tilt angle based on the surface structure of the strawberry. By analyzing the segmentation results, we first identified the stem and tip of the strawberry. Next, we computed the local convex points and fitted a plane to these points, enabling the calculation of the tilt angle relative to the vertical direction.

In the point cloud representation, the tilt angle of a strawberry was calculated as follows: let $W$ represent the point cloud set of the strawberry captured by the camera. The point cloud subsets corresponding to the stem set ($S$) and the tip set ($T$) were utilized to determine the middle belly point cloud set ($B$). Beginning with an initial point ($p_{0}$) in the belly region, an iterative search radius was defined. Within this radius ($r$), the 30 nearest points from the subset $W_{r} = \{p_{0}, \dots, p_{29} \}$ were selected to construct a convex hull. The slope of this convex hull was computed, and the curvature of the nearest convex hull points was iteratively evaluated. The point with the lowest curvature was identified as the central point, representing the local convex apex $p_{h}$ of the strawberry’s belly.

To estimate the tilt angle, this local convex point was connected to a point within the tip point cloud. A random sample of 100 points from $W_{r}$ was selected, and a plane was fitted using the least squares method. The tilt angle of the fitted plane was then determined and used to compute the tilt angle of the strawberry’s central axis relative to the vertical direction. Finally, this angle was used to rectify the tilted strawberry's frontal projection, ensuring an accurate representation.

To estimate strawberry mass, we first established a mapping between the RGB-D data and the fruit mass. After calculating the frontal projection area of the strawberry, the next step was to determine the relationship between the true projection area and the volume. However, since the true frontal projection area and volume did not generally exhibit a direct linear relationship, polynomial regression was used to model this non-linear relationship. The formula is as follows:

\begin{equation}
y = \beta_0 + \beta_1 x + \beta_2 x^2 + \beta_3 x^3 + \dots + \beta_n x^n + \epsilon
\end{equation}

This regression model, combined with the average density of ripe strawberries, enabled the final mass estimation used for grading. This process served as the foundation for strawberry mass estimation under ideal, occlusion-free conditions. The primary source of deviation arose from the lack of corresponding depth information in the RGB images restored by CycleGAN. To mitigate this issue, we supplemented the missing depth data using the average depth of isolated strawberries. In occluded scenarios, the CycleGAN network was integrated to reconstruct the occluded portions of the strawberry, enabling a more accurate mass estimation.

\section{Results and Discussion}

\subsection{Results of strawberry segmentation}

\begin{figure}[htp]
    \centering
    \includegraphics[width=8cm]{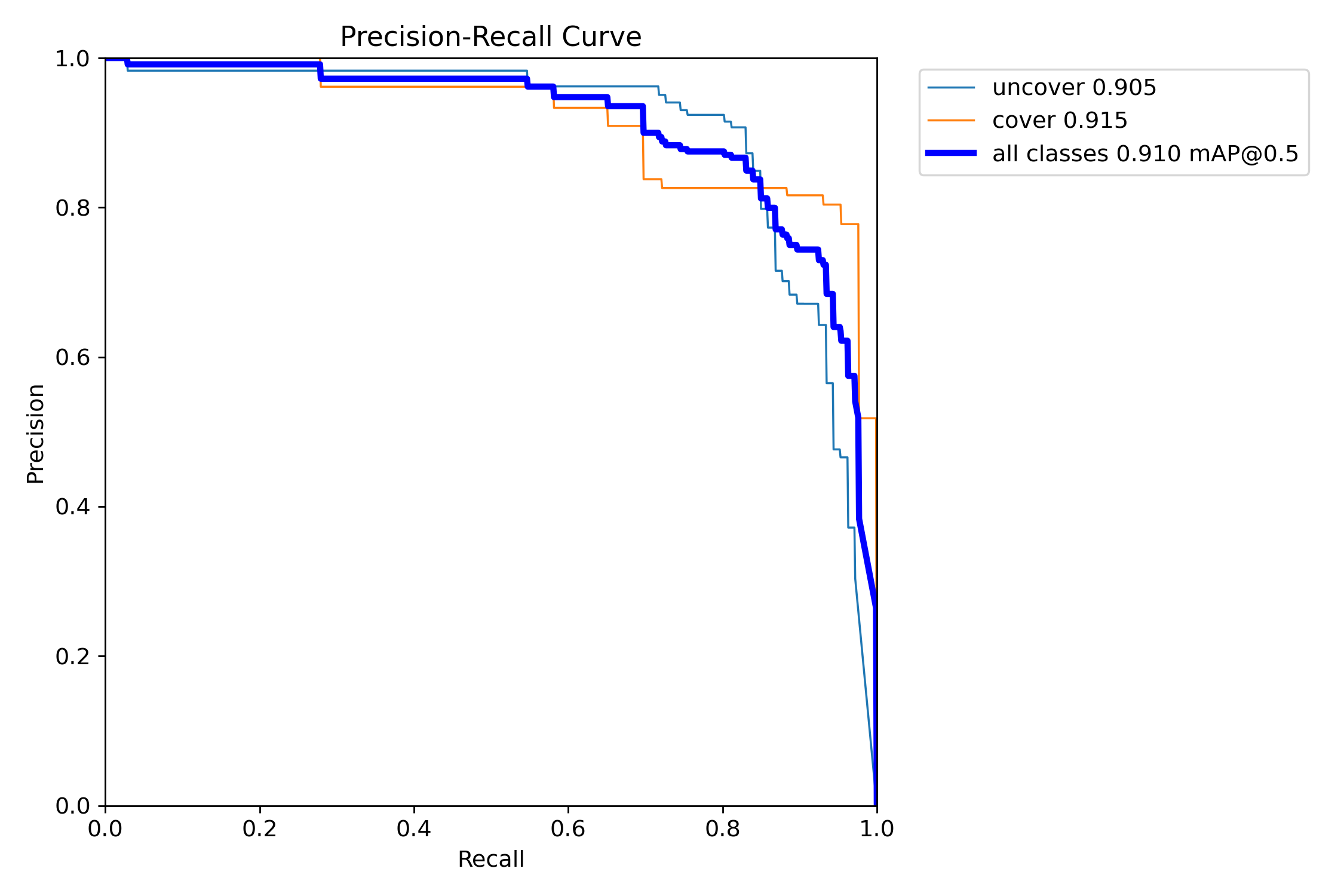}
    \caption{\textbf{Precision–Recall (PR) curves for segmentation under occlusion:} The PR curves illustrate the segmentation performance of YOLOv8-Seg on covered (orange), uncovered (light blue), and all samples (blue bold). The model achieved a mean average precision (mAP@0.5) of 0.910 across all classes, with slightly higher performance on covered samples (0.915) than uncovered ones (0.905), demonstrating robustness against occlusions.}
    \label{fig:yolo}
\end{figure}

YOLOv8-Seg was trained on an Nvidia RTX 4060 GPU using the Adam optimizer with an initial learning rate of 0.01. During training, the model gradually converged after 100 epochs, ultimately achieving a mean average precision (mAP@0.5) of 0.91 on the validation set, as shown in Fig.~\ref{fig:yolo}. The experimental results demonstrated that this method could accurately segment strawberry morphology and maintain high detection accuracy, even in the presence of complex backgrounds and overlapping occlusions. This performance provided strong technical support for subsequent morphological recovery and strawberry mass estimation.

\subsection{Results of occlusion recovery model}

To fully demonstrate the practicality and efficiency of the shape completion method, this study compared the CycleGAN model with the LaMa model, a deep learning approach designed for image restoration in large occluded areas. LaMa achieved a full image-wide receptive field by incorporating fast Fourier convolution (FFC), which enhanced its ability to capture global information efficiently and delivered outstanding performance in restoring large missing regions. For the comparative analysis, we adopted three evaluation metrics: pixel occupancy area, IoU, and inference time. Among these, the pixel occupancy area was defined as the ratio of the pixel area of the restored region to that of the original missing region. The closer this value was to 1, the better the restoration effect.

\begin{figure}[htp]
    \centering
    \includegraphics[width=8cm]{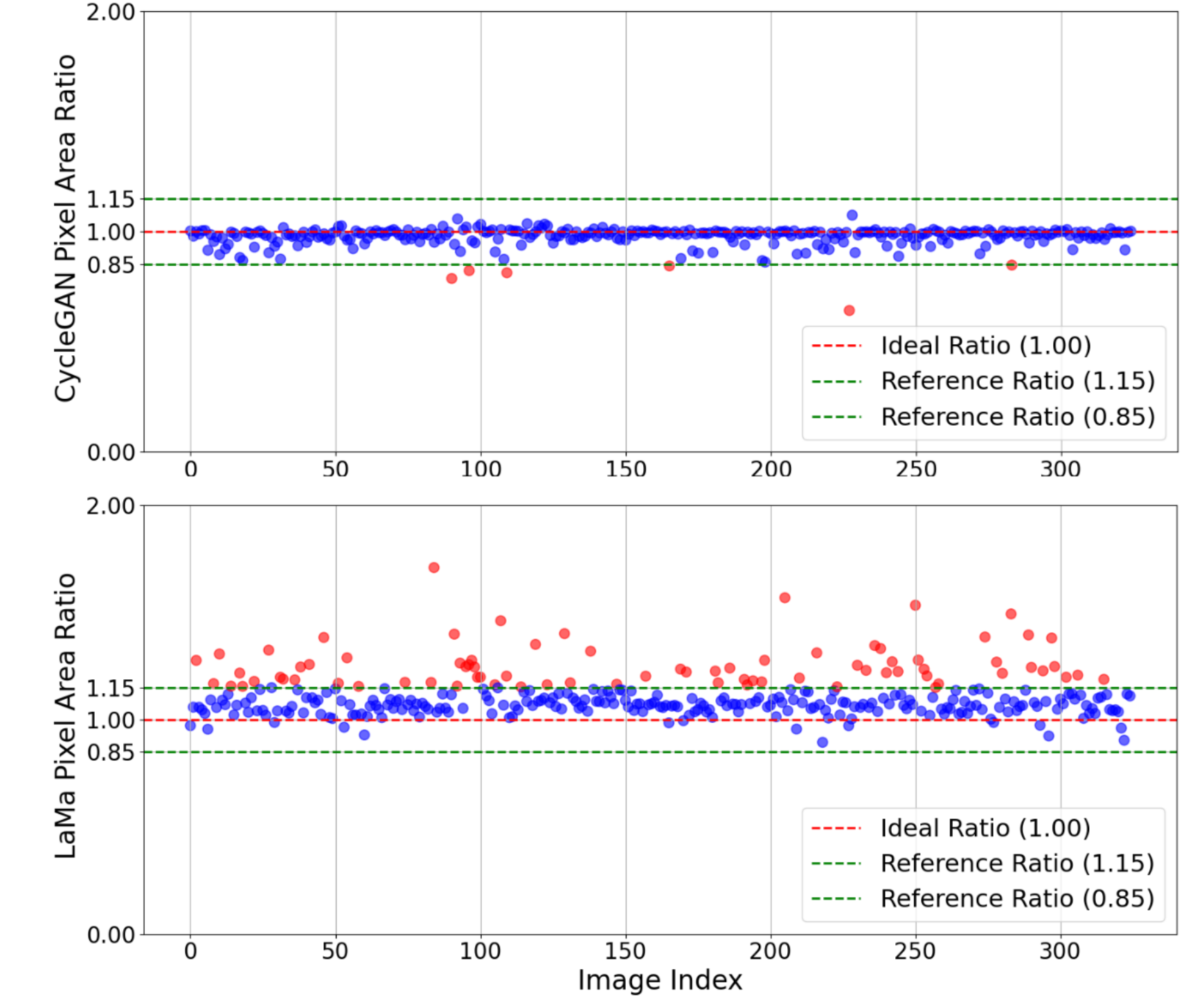}
    \caption{\textbf{Scatter plots of pixel area ratios for CycleGAN and LaMa:} Each dot represents the ratio between the pixel area of the restored fruit and the ground truth across 325 test images. The red dashed line denotes the ideal ratio (1.00), while green dashed lines mark acceptable reference bounds (0.85 and 1.15). The top plot shows that CycleGAN exhibits a tighter distribution around the ideal ratio, indicating better area consistency. In contrast, the bottom plot shows that LaMa yields more outliers and higher variance, suggesting reduced restoration stability.}
    \label{fig:pixel}
\end{figure}

The comparison results are shown in Fig.~\ref{fig:pixel}. The upper plot shows the results of CycleGAN, while the lower plot shows the results of LaMa. The red dots indicate cases that fall outside the range of $1 \pm 0.15$. From the figure, it can be observed that CycleGAN's results are overall denser, with fewer red markers, while LaMa's results are more scattered, with a higher number of red dots. This indicates that CycleGAN achieves better recovery performance compared to LaMa.

\begin{figure}[htp]
    \centering
    \includegraphics[width=9cm]{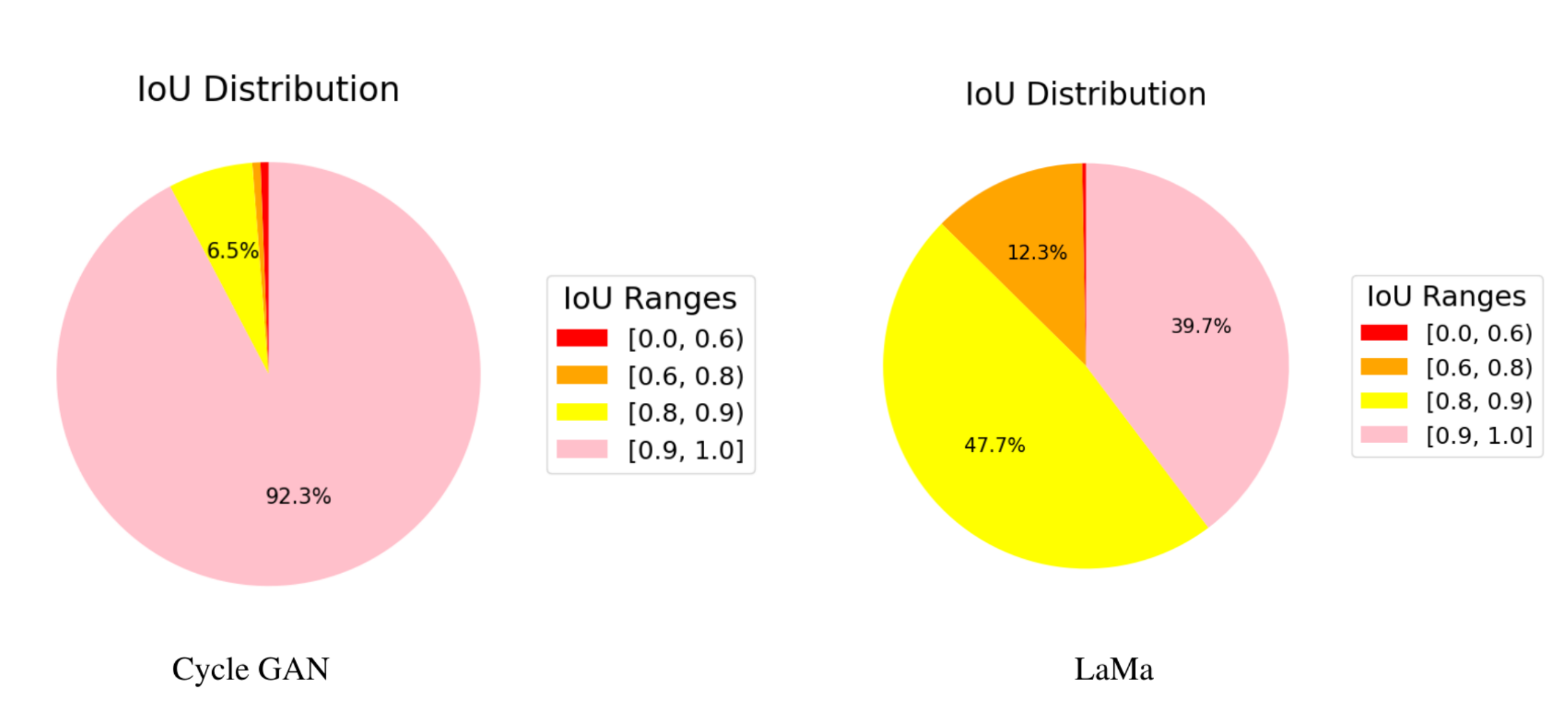}
    \caption{\textbf{IoU distribution comparison between CycleGAN and LaMa:} Pie charts visualize the distribution of IoU values for recovered strawberry contours. The IoU range is divided into four intervals: [0.0, 0.6), [0.6, 0.8), [0.8, 0.9), and [0.9, 1.0]. A higher proportion in the [0.9, 1.0] interval indicates better shape recovery performance. CycleGAN achieves 92.3\% in the highest interval, while LaMa reaches only 39.7\%, confirming the superior contour alignment of CycleGAN.}
    \label{fig:IOU}
\end{figure}

The IoU metric was used to quantify the degree of overlap between the predicted contour and the true contour. It was calculated as the ratio of the intersection area between the predicted and real contours to their union. Specifically, IoU values ranged from 0 to 1, where a value closer to 1 indicated a higher degree of alignment between the predicted and true contours, signifying better restoration performance. As shown in Fig.~\ref{fig:IOU}, 92.3\% of the test cases using CycleGAN achieve IoU values within the high-precision interval of [0.9–1], whereas only 47.7\% of those using LaMa meet this criterion. These results further demonstrate the efficiency and superiority of CycleGAN models in shape completion.

Table~\ref{tab:tab1} presents the detailed performance metrics for both models evaluated on 325 images, including the mean and variance of the pixel occupancy ratio and IoU, as well as the average inference time per frame. The CycleGAN model took approximately 200 ms per frame when running on a GTX 1060. When redeployed to an RTX 4060, the inference time improved to 110 ms, enabling near real-time performance. The mean values indicated the deviation from the ideal value of 1, allowing for a more precise comparison of the models' effectiveness. Overall, CycleGAN outperformed LaMa in both pixel occupancy ratio and IoU. However, it exhibited lower computational efficiency.

\begin{table}[]
    \caption{Quantitative performance comparison between CycleGAN and LaMa. PAR denotes the pixel area ratio.}
    \label{tab:tab1}
    \centering
    \begin{tabular}{cccccc}
        \hline
        Model                & \begin{tabular}[c]{@{}c@{}}Mean\\ (PAR)\end{tabular} & \begin{tabular}[c]{@{}c@{}}$s^2$\\ (PAR)\end{tabular} & \begin{tabular}[c]{@{}c@{}}Mean\\ (IOU)\end{tabular} & \begin{tabular}[c]{@{}c@{}}$s^2$\\ (IOU)\end{tabular} & \begin{tabular}[c]{@{}c@{}}Inference Time\\ (/frame)\end{tabular}  \\ 
        \hline
        LaMa                 & \(1 + 0.112\)                                       & 0.104                                                 & \(1 - 0.13\)                                         & 0.062                                                 & 60 ms  \\ 
        CycleGAN            & \(1 - 0.022\)                                       & 0.022                                                 & \(1 - 0.045\)                                        & 0.083                                                 & 200 ms  \\ 
        \hline
    \end{tabular}
\end{table}

\begin{figure}[htp]
    \centering
    \includegraphics[width=9cm]{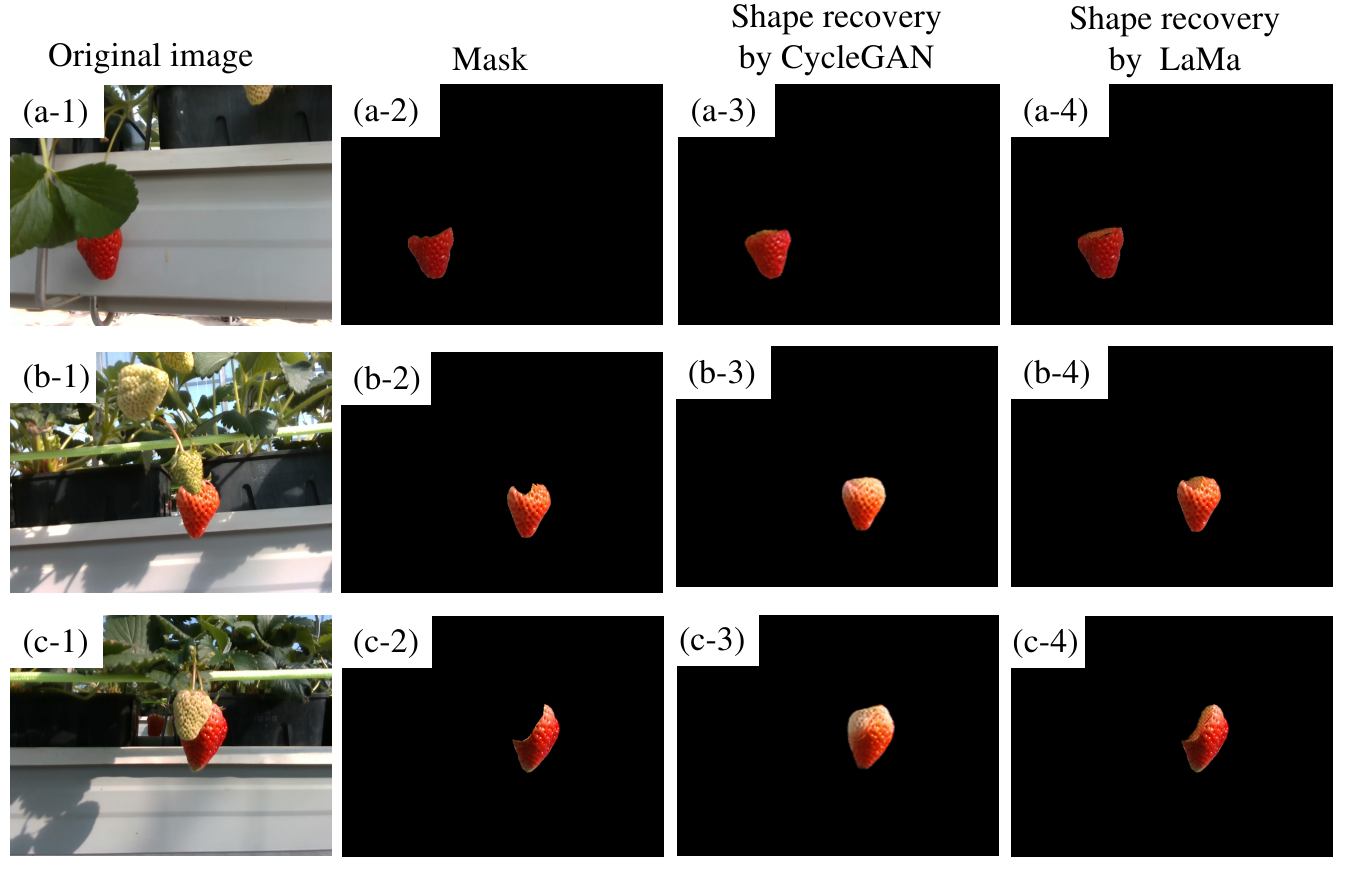}
    \caption{\textbf{Comparison of shape recovery performance:} Visual comparison between CycleGAN and LaMa models in recovering strawberry contours under occlusion. The leftmost column (a-1 to c-1) displays the original RGB images captured in natural field environments. The second column (a-2 to c-2) shows the masked occluded inputs. The third column (a-3 to c-3) presents the contour recovery results using CycleGAN, and the fourth column (a-4 to c-4) shows those obtained by LaMa. CycleGAN demonstrates better contour integrity and symmetry in all cases, especially under large or irregular occlusions.}
    \label{fig:comparison}
\end{figure}

\begin{figure}[htp]
    \centering
    \includegraphics[width=8cm]{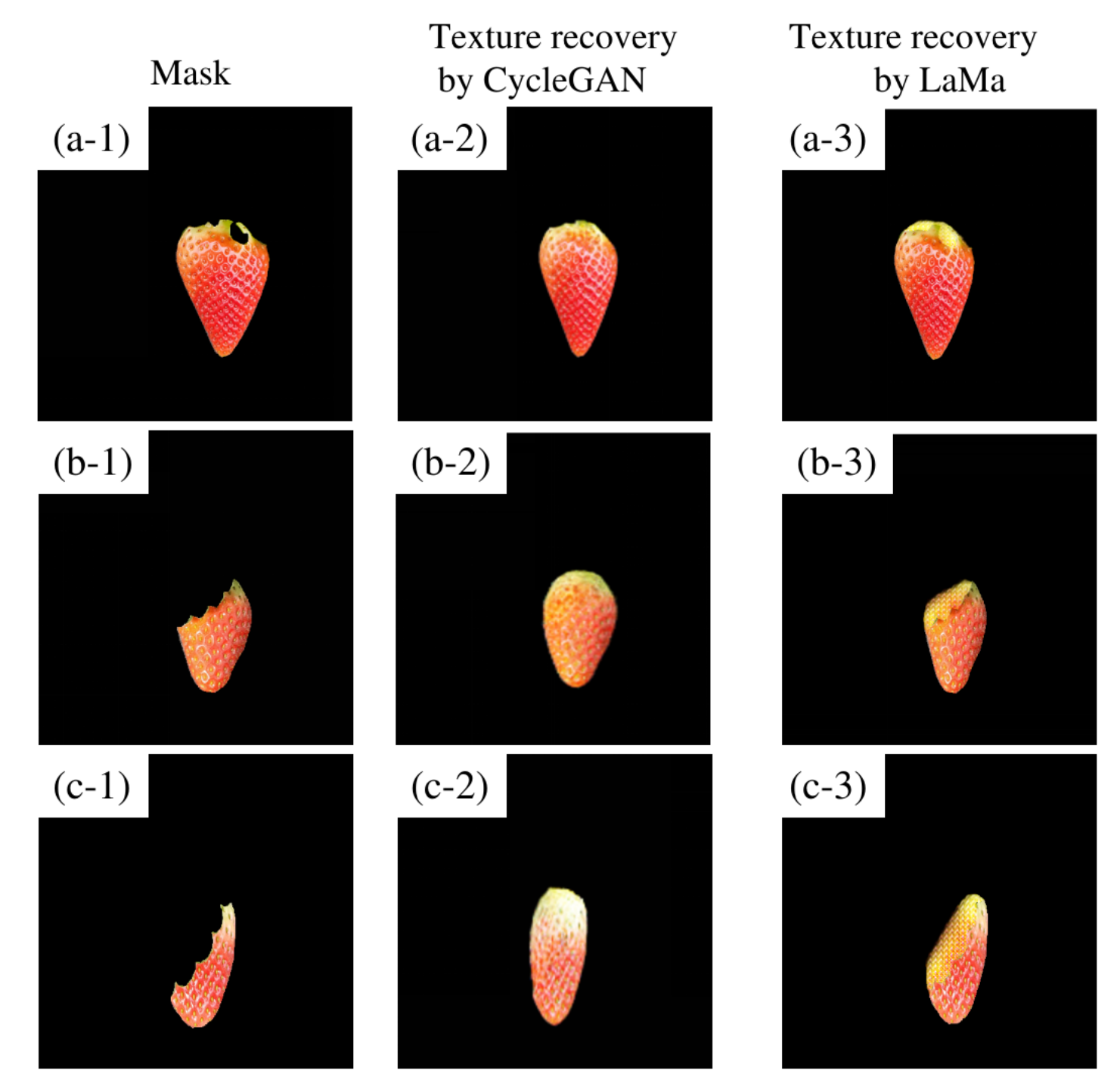}
    \caption{\textbf{Comparison of texture recovery performance:} Visual results of occlusion restoration for three strawberries under different levels of occlusion. The left column (a-1 to c-1) shows masked inputs; the middle column (a-2 to c-2) shows the texture recovery results using CycleGAN; and the right column (a-3 to c-3) shows results from LaMa. Compared to LaMa, CycleGAN produces more visually consistent and complete textures, especially under large missing regions.}
    \label{fig:texture}
\end{figure}

To visually compare the performance of CycleGAN and LaMa in occluded image recovery, Fig.~\ref{fig:comparison} shows the comparative results of the overall contour reconstruction. CycleGAN demonstrated a clear advantage in this regard. As shown in Fig.~\ref{fig:texture}, CycleGAN also performed well in texture restoration, effectively filling occluded regions while maintaining global image consistency. In contrast, the regions restored by LaMa often exhibited inconsistencies with the surrounding textures. However, CycleGAN was constrained by its global reconstruction strategy, and its output resolution was fixed at 256×256 pixels, which may limit the fidelity of fine details.

\subsection{Mass estimation results}

Strawberries, as irregular three-dimensional fruits, exhibit a significant non-linear relationship between their surface projected area and actual volume. This effect is particularly pronounced in smaller fruits, where the volume is highly sensitive to changes in area, whereas volume growth levels off as the area increases. To better characterize this trend, we employed a cubic polynomial regression model. Compared with linear or quadratic models, this approach offered stronger fitting capability and captured the non-linear relationship between area and volume more effectively. The regression formula was derived from measured samples, and its coefficients reflected the empirical correlation between area and volume, enabling efficient estimation of strawberry volume.

To assess the accuracy of the algorithm, a series of comparative experiments were conducted. Field samples were collected to record the maximum vertical cross-sectional area, tilt angle, volume, and mass of the strawberries. The cross-sectional area was measured manually through physical slicing; the tilt angle was obtained using a smartphone-based angle measurement app; the volume was calculated via Archimedes' water displacement method; and the mass was measured with an electronic balance. The measured data points and corresponding fitted curves are shown in Fig.~\ref{fig:reg}.

In addition, we fitted a cubic polynomial regression model to describe the relationship between the maximum cross-sectional area and the volume of the strawberries:

\begin{equation}
y = -24.9926 + 7.1919 \cdot x - 0.3063 \cdot x^{2} + 0.0052 \cdot x^{3}
\end{equation}

As shown in the figure and the regression equation above, a strong correlation between cross-sectional area and volume was observed, with a coefficient of determination \( R^2 = 0.9037 \), indicating a good model fit.

\begin{figure}[htp]
        
	\centering
	\includegraphics[width=0.5\textwidth]{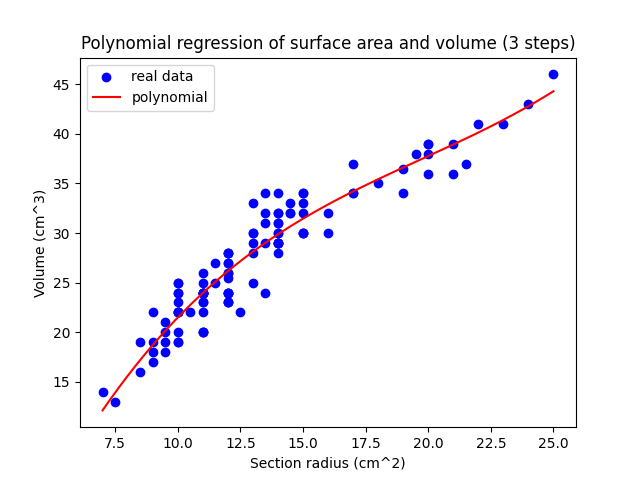}
    \caption{\textbf{Polynomial regression of surface area and volume:} A third-order polynomial regression was used to fit the non-linear relationship between the maximum cross-sectional area and the measured volume of strawberries. Blue dots represent actual measurements, and the red curve shows the fitted polynomial trend.}
	\label{fig:reg}%
\end{figure}

\begin{table}[h]
\caption{Comparison of predicted and actual cross-sectional area, tilt angle, volume, and mass for isolated strawberries.}
        \label{tab:tab2}
\centering
\begin{tabular}{lllll}
\hline
                   & Area  & Angle  & Volume & Mass  \\ \hline
Average error & 4.14\% & 13.71\% & 7.52\%  & 8.11\% \\
Variance           & 1.81\,(cm$^2$)$^2$  & 2.79\,(degree$^2$)$^2$   & 2.04\,(cm$^3$)$^2$   & 1.67\,(g)$^2$  \\ \hline
\end{tabular}
\end{table}

The evaluation results for isolated strawberries are summarized in Table~\ref{tab:tab2}. We compared the actual measured cross-sectional area, tilt angle, and volume of strawberries in the field with the corresponding estimates produced by the proposed method. The mean deviation rate and variance were used as evaluation metrics. 

As shown in the table, for isolated strawberries, the error in the cross-sectional area derived from depth and RGB images remained small. In contrast, the tilt angle estimation exhibited larger errors due to the complexity and variability of field conditions; however, these errors were still within acceptable limits. The volume was estimated from the orthogonal projected area using cubic polynomial regression, yielding an average error of 7.52\%. Based on the average fruit density, the resulting mass estimation had a mean error of 8.11\% and a variance of \( 1.67~\text{g}^2 \).

For occluded strawberries, as presented in Table~\ref{tab:tab3}, the performance of the CycleGAN-based recovery approach showed only a slight increase in estimation error compared to the isolated cases. Although the CycleGAN reconstruction achieved high visual and geometric fidelity, minor deviations introduced during the recovery process contributed to a marginal rise in overall error. Nevertheless, the proposed method maintained strong robustness and reliable performance under occluded conditions.

\begin{table}[h]
 \caption{Prediction errors of area, volume, and mass for shape-completed strawberries.}
 \label{tab:tab3}
 \centering
 \setlength{\tabcolsep}{1pt} 
 \renewcommand{\arraystretch}{0.9} 
 \begin{tabular}{llll}
 \hline
                   & Area       & Volume         & Mass     \\ \hline
Average error & 6.38\%    & 10.07\%      & 10.47\%    \\
Variance          & 2.91\,(cm$^2$)$^2$ & 2.53\,(cm$^3$)$^2$ & 4.76\,(g)$^2$ \\ \hline
\end{tabular}
\end{table}

In addition to quantitative evaluation, we also conducted visual verification of the proposed process in field conditions. As shown in Figure~\ref{fig:reg}: the top row displays original RGB images of strawberries with varying degrees of occlusion; the bottom row shows the final output results, including mass estimation values and grading labels (mass $>$ 30 g is Grade A; $(20,30]$ g is Grade B; $(10,20]$ g is Grade C; $\leq$10 g is Grade D). This case demonstrated the process's ability to perform real-time detection, occlusion repair, and mass prediction under natural growth conditions.

\begin{figure*}[t!]
	\centering
	\includegraphics[width=0.9\linewidth]{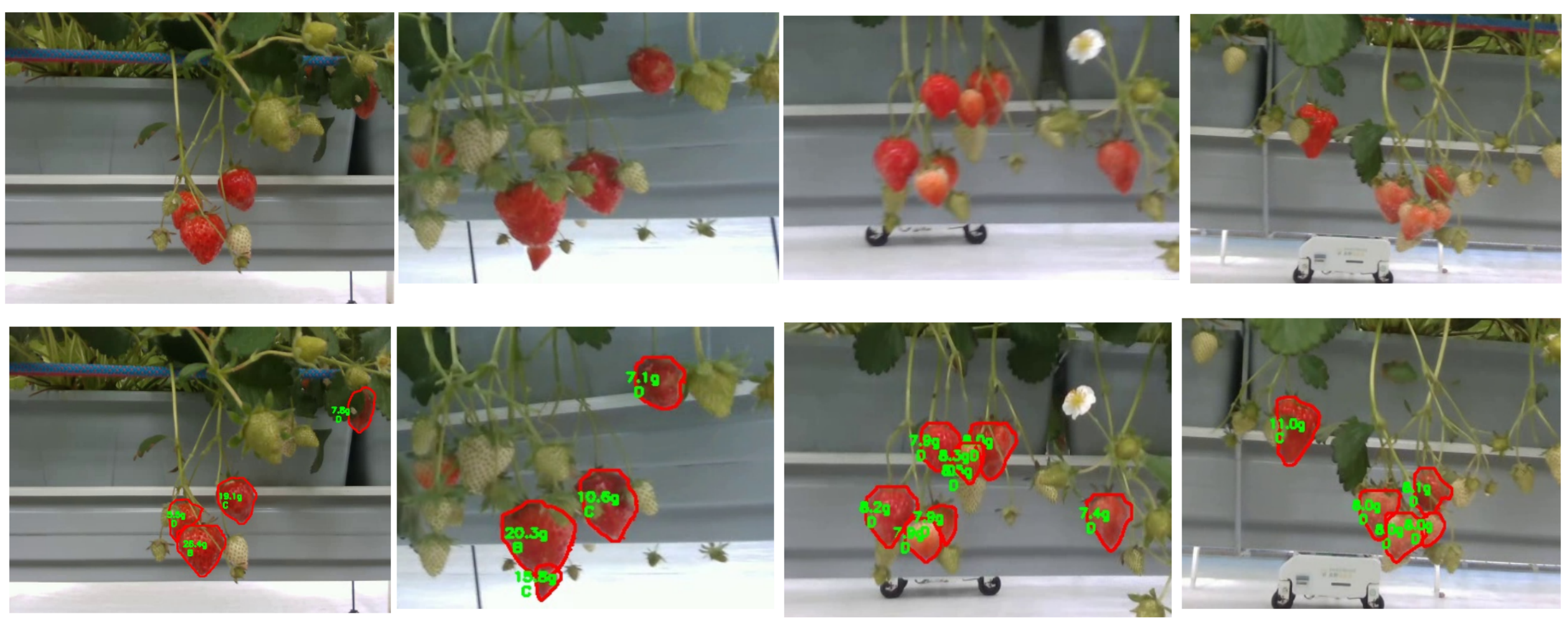}
    \caption{\textbf{Test results under field conditions:} Original RGB images of strawberries under varying occlusion and lighting conditions (top row), final system output showing estimated mass and grade in green text (bottom row).}
	\label{fig:reg}%
\end{figure*}

\section{DISCUSSION}
This method was based on data collected by an RGB-D camera, but under complex lighting conditions, depth maps were susceptible to noise. To address this issue, multi-view or multi-modal data and adaptive repair strategies can be combined in subsequent steps to improve geometric feature retention and estimation accuracy. During image restoration, the input to the CycleGAN network was fixed at 256×256 pixels. Despite normalisation, some strawberry details and textures may still be distorted, thereby affecting restoration mass and the accuracy of subsequent masss estimation. To address this issue, future research could further enhance CycleGAN's detail restoration capabilities and improve the accuracy and robustness of mass prediction by optimising the network structure, introducing higher-resolution inputs, or integrating multi-scale feature fusion. Additionally, although polynomial regression was generally effective, it is highly sensitive to data distribution and may produce significant errors when handling irregularly shaped strawberries. This method also assumed a fixed mapping relationship between mass and volume, without considering density changes caused by factors such as cultivation variety, maturity, and moisture content, which also affected the accuracy of mass estimation. To enhance prediction reliability, future research could explore more robust nonlinear methods.

\section{CONCLUSIONS}
This work presented a vision-based pipeline for online mass estimation of table-top grown strawberries under field conditions, addressing occlusion challenges through CycleGAN-based shape completion and geometry correction informed by pose estimation. By combining YOLOv8-Seg segmentation, unsupervised occlusion recovery, and tilt-adjusted projection analysis, the method achieved a mean mass estimation error below 10.5\% for both isolated and occluded fruits. Key innovations included the adaptation of CycleGAN for agricultural occlusion recovery and a symmetry-based pose estimation algorithm.


\bibliographystyle{unsrt}
\bibliography{conference_101719}

\begin{thebibliography}{10}

\bibitem{b1}
Chiranjivi Neupane, Maisa Pereira, Anand Koirala, and Kerry~B Walsh.
\newblock Fruit sizing in orchard: A review from caliper to machine vision with deep learning.
\newblock {\em Sensors}, 23(8):3868, 2023.

\bibitem{b2}
Naoshi Kondo.
\newblock Automation on fruit and vegetable grading system and food traceability.
\newblock {\em Trends in Food Science \& Technology}, 21(3):145--152, 2010.

\bibitem{xiong2020autonomous}
Ya~Xiong, Yuanyue Ge, Lars Grimstad, and P{\aa}l~J From.
\newblock An autonomous strawberry-harvesting robot: Design, development, integration, and field evaluation.
\newblock {\em Journal of Field Robotics}, 37(2):202--224, 2020.

\bibitem{b011}
Davide Cassanelli, Nicola Lenzini, Luca Ferrari, and Luigi Rovati.
\newblock Partial least squares estimation of crop moisture and density by near-infrared spectroscopy.
\newblock {\em IEEE Transactions on Instrumentation and Measurement}, 70:1--10, 2021.

\bibitem{b012}
Yuanshuo Hao, Faris Rafi~Almay Widagdo, Xin Liu, Ying Quan, Lihu Dong, and Fengri Li.
\newblock Individual tree diameter estimation in small-scale forest inventory using uav laser scanning.
\newblock {\em Remote Sensing}, 13(1):24, 2020.

\bibitem{b013}
Hae-Il Yang, Sung-Gi Min, Ji-Hee Yang, Jong-Bang Eun, and Young-Bae Chung.
\newblock A novel hybrid-view technique for accurate mass estimation of kimchi cabbage using computer vision.
\newblock {\em Journal of Food Engineering}, 378:112126, 2024.

\bibitem{b13}
Juan~C Miranda, Jaume Arn{\'o}, Jordi Gen{\'e}-Mola, Jaume Lordan, Luis As{\'\i}n, and Eduard Gregorio.
\newblock Assessing automatic data processing algorithms for rgb-d cameras to predict fruit size and weight in apples.
\newblock {\em Computers and Electronics in Agriculture}, 214:108302, 2023.

\bibitem{b16}
Aharon Kalantar, Yael Edan, Amit Gur, and Iftach Klapp.
\newblock A deep learning system for single and overall weight estimation of melons using unmanned aerial vehicle images.
\newblock {\em Computers and Electronics in Agriculture}, 178:105748, 2020.

\bibitem{tafuro2022strawberry}
Alessandra Tafuro, Adeayo Adewumi, Soran Parsa, Ghalamzan~E Amir, and Bappaditya Debnath.
\newblock Strawberry picking point localization ripeness and weight estimation.
\newblock In {\em 2022 International conference on robotics and automation (ICRA)}, pages 2295--2302. Ieee, 2022.

\bibitem{huang2024strawberry}
Yanjiang Huang, Jiepeng Liu, and Xianmin Zhang.
\newblock Strawberry weight estimation based on plane-constrained binary division point cloud completion.
\newblock In {\em 2024 IEEE International Conference on Robotics and Automation (ICRA)}, pages 11846--11852. IEEE, 2024.

\bibitem{basak2022non}
Jayanta~Kumar Basak, Bhola Paudel, Na~Eun Kim, Nibas~Chandra Deb, Bolappa~Gamage Kaushalya~Madhavi, and Hyeon~Tae Kim.
\newblock Non-destructive estimation of fruit weight of strawberry using machine learning models.
\newblock {\em Agronomy}, 12(10):2487, 2022.

\bibitem{b016}
Lin~Mar Oo and Nay~Zar Aung.
\newblock A simple and efficient method for automatic strawberry shape and size estimation and classification.
\newblock {\em Biosystems engineering}, 170:96--107, 2018.

\bibitem{Ge_symmetry}
Yuanyue Ge, Ya~Xiong, and Pål~J. From.
\newblock Symmetry-based 3d shape completion for fruit localisation for harvesting robots.
\newblock {\em Biosystems Engineering}, 197:188--202, 2020.

\bibitem{b22}
Yongsheng Yu, Libo Zhang, Heng Fan, and Tiejian Luo.
\newblock High-fidelity image inpainting with gan inversion.
\newblock In {\em European Conference on Computer Vision}, pages 242--258. Springer, 2022.

\bibitem{b23}
Ankan Dash, Jingyi Gu, and Guiling Wang.
\newblock Hi-gan: Hierarchical inpainting gan with auxiliary inputs for combined rgb and depth inpainting.
\newblock {\em arXiv preprint arXiv:2402.10334}, 2024.

\bibitem{b24}
Andranik Sargsyan, Shant Navasardyan, Xingqian Xu, and Humphrey Shi.
\newblock Mi-gan: A simple baseline for image inpainting on mobile devices.
\newblock In {\em Proceedings of the IEEE/CVF International Conference on Computer Vision}, pages 7335--7345, 2023.

\bibitem{10944053}
Jiahui Li, Pourya Shamsolmoali, Yue Lu, and Masoumeh Zareapoor.
\newblock Shapemorph: 3d shape completion via blockwise discrete diffusion.
\newblock pages 2818--2827, 2025.

\bibitem{chu2023diffcompletediffusionbasedgenerative3d}
Ruihang Chu, Enze Xie, Shentong Mo, Zhenguo Li, Matthias Nießner, Chi-Wing Fu, and Jiaya Jia.
\newblock Diffcomplete: Diffusion-based generative 3d shape completion.
\newblock 2023.

\bibitem{b8x}
Jun-Yan Zhu, Taesung Park, Phillip Isola, and Alexei~A Efros.
\newblock Unpaired image-to-image translation using cycle-consistent adversarial networks.
\newblock In {\em Proceedings of the IEEE international conference on computer vision}, pages 2223--2232, 2017.

\bibitem{b017}
Kaiming He, Georgia Gkioxari, Piotr Doll{\'a}r, and Ross Girshick.
\newblock Mask r-cnn.
\newblock In {\em Proceedings of the IEEE international conference on computer vision}, pages 2961--2969, 2017.

\bibitem{b018}
Olaf Ronneberger, Philipp Fischer, and Thomas Brox.
\newblock U-net: Convolutional networks for biomedical image segmentation.
\newblock In {\em Medical image computing and computer-assisted intervention--MICCAI 2015: 18th international conference, Munich, Germany, October 5-9, 2015, proceedings, part III 18}, pages 234--241. Springer, 2015.

\bibitem{b019}
Chien-Yao Wang, Hong-Yuan~Mark Liao, Yueh-Hua Wu, Ping-Yang Chen, Jun-Wei Hsieh, and I-Hau Yeh.
\newblock Cspnet: A new backbone that can enhance learning capability of cnn.
\newblock In {\em Proceedings of the IEEE/CVF conference on computer vision and pattern recognition workshops}, pages 390--391, 2020.

\end{thebibliography}



\end{document}